\documentclass[twoside,11pt]{article}
\usepackage{pgm}

\usepackage[utf8]{inputenc}
\inputencoding{utf8}

\usepackage{hyperref,color,soul,booktabs}
\setulcolor{blue}
\newcommand{\BibTeX}{\textsc{B\kern-0.1emi\kern-0.017emb}\kern-0.15em\TeX}

\usepackage{amsmath, bm}
\usepackage{tikz}
\usetikzlibrary{arrows, calc, matrix}
\usepackage{enumitem}
\usepackage{subcaption}
\usepackage{algorithm, algpseudocode}
\usepackage{array, colortbl}
\usepackage{makecell}

\newcommand{\nobs}{n}

\newcommand{\bfzero}{\mathbf{0}} 
\newcommand{\cov}{\mathbf{\Sigma}} 
\newcommand{\cove}{\Sigma} 
\newcommand{\scm}{\mathbf{S}} 
\newcommand{\scme}{S} 
\newcommand{\cor}{\mathbf{R}} 
\newcommand{\core}{R} 
\newcommand{\scor}{\mathbf{C}} 
\newcommand{\score}{C} 
\newcommand{\scl}{\mathbf{\Psi}} 
\newcommand{\var}{{\textit{X}}} 
\newcommand{\varv}{{\mathbf{x}}} 
\newcommand{\data}{{\mathbf{D}}} 
\newcommand{\model}{{\mathcal{M}}} 
\newcommand{\given}{\:|\:} 
\newcommand{\Bf}{\mathcal{B}} 
\newcommand{\idm}{\mathbf{I}} 
\newcommand{\imod}{{\mathcal{I}}} 

\newcommand{\Wish}[1]{\mathcal{W}_{#1}} 
\newcommand{\iWish}[1]{\Wish{#1}^{-1}} 
\newcommand{\pWish}{{\mathcal{PW}}} 
\newcommand{\ipWish}{{\pWish^{-1}}} 

\newcommand{\tr}[1]{\textrm{tr}\left(#1\right)}
\newcommand{\diff}[1]{\:\textrm{d}{#1}\:}
\newcommand{\half}[1]{{\frac{#1}{2}}}
\newcommand{\trp}{\intercal} 

\newcolumntype{C}[1]{>{\centering\arraybackslash}m{#1}}

\tikzset{
	square matrix/.style={
		matrix of nodes,
		nodes in empty cells,
		column sep=-\pgflinewidth, row sep=-\pgflinewidth,
		nodes={draw=black,
			minimum height=#1,
			anchor=center,
			text width=#1,
			align=center,
			inner sep=0pt,
			fill = gray,
		},
	},
	square matrix/.default=3ex,
	myset/.style args = {(#1, #2)}{%
		row #1 column #2/.style={
			nodes = {
				draw = black, 
				fill = none, 
				text = black,
				inner sep=0pt, 
				font={\Large 0}
			}
		}
	},
}

\makeatletter
\newcommand*{\indep}{%
	\mathbin{%
		\mathpalette{\@indep}{}%
	}%
}
\newcommand*{\nindep}{%
	\mathbin{
		\mathpalette{\@indep}{\not}
	}%
}
\newcommand*{\@indep}[2]{%
	\sbox0{$#1\perp\m@th$}
	\sbox2{$#1=$}
	\sbox4{$#1\vcenter{}$}
	\rlap{\copy0}
	\dimen@=\dimexpr\ht2-\ht4-.2pt\relax
	\kern\dimen@
	{#2}%
	\kern\dimen@
	\copy0 
}
\makeatother

\ShortHeadings{A Bayesian Approach for Inferring Local Causal Structure in Gene Regulatory Networks}{Bucur et al.}

\begin{document}

\title{A Bayesian Approach for Inferring Local Causal Structure in Gene Regulatory Networks}

\author{\Name{Ioan Gabriel Bucur} \Email{g.bucur@cs.ru.nl} \and
	\Name{Tom van Bussel} \Email{tom.van.bussel@student.ru.nl} \and
	\Name{Tom Claassen} \Email{tomc@cs.ru.nl} \and 
	\Name{Tom Heskes} \Email{t.heskes@science.ru.nl} \\
	\addr Faculty of Science, Radboud University, Postbus 9010, 6500GL Nijmegen, The Netherlands}

\maketitle

\begin{abstract}

Gene regulatory networks play a crucial role in controlling an organism's biological processes, which is why there is significant interest in developing computational methods that are able to extract their structure from high-throughput genetic data. A typical approach consists of a series of conditional independence tests on the covariance structure meant to progressively reduce the space of possible causal models. We propose a novel efficient Bayesian method for discovering the local causal relationships among triplets of (normally distributed) variables. In our approach, we score the patterns in the covariance matrix in one go and we incorporate the available background knowledge in the form of priors over causal structures. Our method is flexible in the sense that it allows for different types of causal structures and assumptions. We apply the approach to the task of inferring gene regulatory networks by learning regulatory relationships between gene expression levels. We show that our algorithm produces stable and conservative posterior probability estimates over local causal structures that can be used to derive an honest ranking of the most meaningful regulatory relationships. We demonstrate the stability and efficacy of our method both on simulated data and on real-world data from an experiment on yeast.

\end{abstract}


\section{Introduction}

Gene regulatory networks (GRNs) play a crucial role in controlling an organism's biological processes, such as cell differentiation and metabolism. If we knew the structure of a GRN, we would be able to intervene in the developmental process of the organism, for instance by targeting a specific gene with drugs. In recent years, researchers have developed a number of methods for inferring regulatory relationships from data on gene expression, the process by which genetic instructions are used to synthesize gene products such as proteins. Gene regulatory relationships are inherently causal: we can manipulate the expression level of one gene (the `cause') to regulate that of another gene (the `effect'). Because of this, many GRN inference algorithms such as `Trigger'~\citep{chen_harnessing_2007}, `CIT'~\citep{millstein_disentangling_2009} or `CMST'~\cite{neto_modeling_2013} are aimed at finding promising causal regulatory relationships among genes.

An efficient way to derive causal relationships from observational data, which results in clear and easily interpretable output, is to find local causal patterns in the data. The \textit{local causal discovery} (LCD) algorithm~\citep{cooper_simple_1997} makes use of a combination of observational data and background knowledge when searching for unconfounded causal relationships among triplets of variables. Trigger is also designed to search for this LCD pattern in the data, using the background knowledge that genetic information is randomized at birth, before any other measurements can be made. \cite{mani_theoretical_2006}, on the other hand, divide the causal discovery task into identifying so-called Y structures on subsets of four variables. The Y structure is the smallest pattern containing an unconfounded causal relationship that can be learned solely from observational data in the presence of latent variables. 

A key feature of Trigger is that it can estimate the probability of causal regulatory relationships, while controlling for the false discovery rate~\citep{chen_harnessing_2007}. The algorithm consists of a series of likelihood ratio tests for regression coefficients that are translated into statements about conditional (in)dependence, which are then used to identify the presence of the LCD pattern. Testing whether regression coefficients are significantly different from zero essentially boils down to testing whether partial correlation coefficients are significantly different from zero~\citep{nicolai_meinshausen_high-dimensional_2006}, which means that all the information needed for the tests lies in the covariance structure.

We propose a Bayesian approach for local causal discovery on triplets of (normally distributed) variables that makes use of the information in the covariance structure. With our method, we directly score patterns in the data by computing posterior probabilities over all possible three-dimensional covariance structures in one go, with the end goal of identifying plausible causal relationships. This provides a stable, efficient and elegant way of expressing the uncertainty in the underlying local causal structure, even in the presence of latent variables. Moreover, it is straightforward to incorporate background knowledge in the form of priors on causal structures. We show how we can plug in our method into an algorithm that searches for local causal patterns in a GRN and outputs a well-calibrated and reliable ranking of the most likely causal regulatory relationships.

The rest of the paper is organized as follows. In Section~\ref{sec:bkg}, we introduce some standard background notation and terminology. In Section~\ref{sec:bfcs}, we describe our Bayesian approach for inferring the covariance pattern of a three-dimensional Gaussian random vector. By defining simple priors, we then derive the posterior probabilities of local causal structures given the data. In Section~\ref{sec:experiments}, we present the results of applying our method on simulated and real-world data. We conclude by discussing advantages and disadvantages of our approach in Section~\ref{sec:discussion}.

\section{Background} \label{sec:bkg}

Causal structures can be represented by directed graphs, where the nodes in the graph represent (random) variables and the edges between nodes represent causal relationships. \textit{Maximal ancestral graphs} (MAGs) encode conditional independence information and causal relationships in the presence of latent variables and selection bias~\citep{richardson_ancestral_2002}. We refer to MAGs without undirected edges as \textit{directed maximal ancestral graphs} (DMAGs). DMAGs are closed under marginalization, which means they preserve the conditional independence information in the presence of latent variables.

Two causal structures are \textit{Markov (independence) equivalent} if they imply the same conditional independence statements. The \textit{Markov equivalence class} of a MAG (or DMAG) is represented by a \text{partial ancestral graph} (PAG), which displays all the edge marks (arrowhead or tail) shared by all members in the class and displays circles for those marks that are not common among all members. In this work, we will consider two types of graphs: \textit{directed acyclic graphs} and \textit{directed maximal ancestral graphs}. However, the results presented can be applied to any causal graph structure.

A \textit{(conditional) independence model} $\imod$ over a finite set of variables $V$ is a set of triples $\left<X, Y \given Z\right>$, called \textit{(conditional) independence statements}, where $X, Y, Z$ are disjoint subsets of $V$ and $Z$ may be empty~\citep{studeny_probabilistic_2006}. We can induce a (probabilistic) independence model over a probability distribution $P \in \mathcal{P}$ by letting:
$$\left<A, B \given C\right> \in \imod(P) \iff A \indep B \given C \textrm{ w.r.t. } P.$$

The conditional independence model induced by a multivariate Gaussian distribution is a \textit{compositional graphoid}~\citep{sadeghi_markov_2014}, which means that it satisfies the \textit{graphoid axioms} and the \textit{composition} property. Because of this, there is a one-to-one correspondence between the conditional independence models that can be induced by a multivariate Gaussian and the Markov equivalence classes of a causal graph structure.

\section{Bayes Factors of Covariance Structures (BFCS)}  \label{sec:bfcs}

\begin{figure}[!htb]
	\centering
	\begin{tabular}{c | C{.46\linewidth} | C{.12\linewidth} C{.12\linewidth}}
		Model & Markov Equivalence Class (PAG) & Covariance Matrix & Precision Matrix \\
		\hline \hline
		&&& \\
		\makecell{`$\var_1 \nindep \var_2 \nindep \var_3$' \\ (Full)} &
		\begin{tikzpicture}[shorten >=1pt, auto, node distance = 2cm, thick]
		\tikzset{vertex/.style = {shape = circle, draw, minimum size=0.5cm}}
		\tikzset{edge/.style = {->,> = latex'}}
		
		\node[vertex] (1) at (0, 0) {$\var_1$};
		\node[vertex] (2) at (3, 0) {$\var_2$};
		\node[vertex] (3) at (6, 0) {$\var_3$};
		
		\draw[edge, o-o] (1) to (2);
		\draw[edge, o-o] (2) to (3);
		\draw[edge, o-o] (1) to[bend right] (3);
		\end{tikzpicture} &
		\begin{tikzpicture}
		\matrix (cor) [square matrix, myset/.list={}] {
			&   &   \\   &   &   \\   &   &   \\
		};
		\end{tikzpicture} &
		\begin{tikzpicture}
		\matrix (prc) [square matrix, myset/.list={}] {
			&   &   \\   &   &   \\   &   &   \\
		};
		\end{tikzpicture} \\
		\makecell{`$\var_1 \indep \var_3$' \\ (Acausal)} &
		\begin{tikzpicture}[shorten >=1pt, auto, node distance = 2cm, thick]
		\tikzset{vertex/.style = {shape = circle, draw, minimum size=0.5cm}}
		\tikzset{edge/.style = {->,> = latex'}}
		
		\node[vertex] (1) at (0, 0) {$\var_1$};
		\node[vertex] (2) at (3, 0) {$\var_2$};
		\node[vertex] (3) at (6, 0) {$\var_3$};
		
		\draw[edge, o->] (1) to (2);
		\draw[edge, <-o] (2) to (3);
		\end{tikzpicture} &
		\begin{tikzpicture}
		\matrix (cor) [square matrix, myset/.list={(1, 3), (3, 1)}] {
			&   &   \\   &   &   \\   &   &   \\
		};
		\end{tikzpicture} &
		\begin{tikzpicture}
		\matrix (prc) [square matrix, myset/.list={}] {
			&   &   \\   &   &   \\   &   &   \\
		};
		\end{tikzpicture} \\
		\makecell{`$\var_1 \indep \var_3 \given \var_2$' \\ (Causal)} & 
		\begin{tikzpicture}[shorten >=1pt, auto, node distance = 2cm, thick]
		\tikzset{vertex/.style = {shape = circle, draw, minimum size=0.5cm}}
		\tikzset{edge/.style = {->,> = latex'}}
		
		\node[vertex] (1) at (0, 0) {$\var_1$};
		\node[vertex] (2) at (3, 0) {$\var_2$};
		\node[vertex] (3) at (6, 0) {$\var_3$};
		
		\draw[edge, o-o] (1) to (2);
		\draw[edge, o-o] (2) to (3);
		\end{tikzpicture} &
		\begin{tikzpicture}
		\matrix (cor) [square matrix, myset/.list={}] {
			&   &   \\   &   &   \\   &   &   \\
		};
		\end{tikzpicture} &
		\begin{tikzpicture}
		\matrix (prc) [square matrix, myset/.list={(1, 3), (3, 1)}] {
			&   &   \\   &   &   \\   &   &   \\
		};
		\end{tikzpicture} \\
		\makecell{`$(\var_1, \var_3) \indep \var_2$' \\ (Independent)} &
		\begin{tikzpicture}[shorten >=1pt, auto, node distance = 2cm, thick]
		\tikzset{vertex/.style = {shape = circle, draw, minimum size=0.5cm}}
		\tikzset{edge/.style = {->,> = latex'}}
		
		\node[vertex] (1) at (0, 0) {$\var_1$};
		\node[vertex] (2) at (3, 0) {$\var_2$};
		\node[vertex] (3) at (6, 0) {$\var_3$};
		
		\draw[edge, o-o, bend right] (1) to (3);
		\end{tikzpicture} &
		\begin{tikzpicture}
		\matrix (cor) [square matrix, myset/.list={(1, 2), (2, 3), (2, 1), (3, 2)}] {
			&   &   \\   &   &   \\   &   &   \\
		};
		\end{tikzpicture} &
		\begin{tikzpicture}
		\matrix (prc) [square matrix, myset/.list={(1, 2), (2, 3), (2, 1), (3, 2)}] {
			&   &   \\   &   &   \\   &   &   \\
		};
		\end{tikzpicture} \\
		\makecell{`$\var_1 \indep \var_2 \indep \var_3$' \\ (Empty)} &
		\begin{tikzpicture}[shorten >=1pt, auto, node distance = 2cm, thick]
		\tikzset{vertex/.style = {shape = circle, draw, minimum size=0.5cm}}
		\tikzset{edge/.style = {->,> = latex'}}
		
		\node[vertex] (1) at (0, 0) {$\var_1$};
		\node[vertex] (2) at (3, 0) {$\var_2$};
		\node[vertex] (3) at (6, 0) {$\var_3$};
		\end{tikzpicture} &
		\begin{tikzpicture}
		\matrix (cor) [square matrix, myset/.list={(1, 2), (1, 3), (2, 3), (2, 1), (3, 1), (3, 2)}] {
			&   &   \\   &   &   \\   &   &   \\
		};
		\end{tikzpicture} &
		\begin{tikzpicture}
		\matrix (prc) [square matrix, myset/.list={(1, 2), (1, 3), (2, 3), (2, 1), (3, 1), (3, 2)}] {
			&   &   \\   &   &   \\   &   &   \\
		};
		\end{tikzpicture} \\
	\end{tabular}
	\caption{Overview of the five canonical independence patterns between three variables, depicting the equivalence between causal models, conditional independences, and covariance structures} \label{fig:equivalence}
\end{figure}

We are interested in inferring the local covariance structure from observational data, assuming the data follows a (latent) Gaussian model. We will be working with triplets of variables. With finite data, we can never be sure about the true covariance structure underlying the data. Hence, we prefer to work with probability distributions over covariance matrices. For a general three-dimensional covariance matrix $\cov$, the likelihood reads: $$ p(\data \given \cov) = (2\pi)^{-\frac{3n}{2}} |\cov|^{-\frac{n}{2}} \exp\left[-\frac{1}{2} \tr{\scm \cov^{-1}} \right],$$ where $\scm = \data^\trp \data$ is the scatter matrix. 

Under the Gaussianity assumption, there is a one-to-one correspondence between the constraints in the covariance matrix and the conditional independences among the variables. There are five specific canonical patterns to consider, which are depicted in Figure~\ref{fig:equivalence}. The `full' and `empty' covariance patterns are self-explanatory. We call `independent' the pattern occurring when one variable is independent of the other two. We call the pattern on the second row `acausal' because $\var_2$ cannot cause $\var_1$ or $\var_3$ if conditioning upon $\var_2$ turns a conditional independence between $\var_1$ and $\var_3$ into a conditional dependence. We call the pattern on the third row `causal' because $\var_2$ either causes $\var_1$ or $\var_3$ if conditioning upon $\var_2$ turns a conditional dependence between $\var_1$ and $\var_3$ into a conditional independence~\citep{claassen_logical_2011}. The five patterns translate into eleven distinct covariance structures when considering all permutations of three variables. These are the only possible covariance structures on three variables, since the conditional independence model induced by a multivariate Gaussian is a compositional graphoid~\citep{sadeghi_markov_2014}.

Our goal is to compute the posterior probability of each of the possible conditional independence models given the data. We denote by $\mathcal{J} = \{\model_0, \model_1, ..., \model_{10}\}$ the set of all possible conditional independence models. The model evidence is then, for $\model_j \in \mathcal{J}$: $$p(\data \given \model_j) = \int \diff \cov p(\data \given \cov) p(\cov \given \model_j).$$

To facilitate computation, we derive the Bayes factors of each conditional independence model ($\model_j$) compared to a reference independence model ($\model_0$): 

\begin{equation*}
	\Bf_j = \frac{p(\data \given \model_j)}{p(\data \given \model_0)} = \frac{\int \diff{\cov} p(\data \given \cov) p(\cov \given \model_j)}{\int \diff{\cov} p(\data \given \cov) p(\cov \given \model_0)}.
\end{equation*}

As we shall see in Subsection~\ref{ssec:derivation}, many terms will cancel out, making the resulting ratios much simpler to compute (see for example Equation~\ref{eqn:bf_indep}). Finally, we arrive at the posterior probabilities:
\begin{equation}
p(\model_j \given \data) = \frac{p(\data \given \model_j) \cdot p(\model_j)}{\sum_j p(\data \given \model_j) \cdot p(\model_j)} = \frac{\Bf_j \cdot p(\model_j)}{\sum_j \Bf_j \cdot p(\model_j)}.  \label{eqn:post_ci_model}
\end{equation}

\subsection{Choosing the Prior on Covariance Matrices} \label{ssec:choosing_prior_covm}

We consider the inverse Wishart distribution for three-dimensional covariance matrices, which is parameterized by the positive definite scale matrix $\scl$ and the number of degrees of freedom $\nu$: $$ \cov \sim \iWish{3} (\scl, \nu); \quad  p(\cov) = \frac{|\scl|^{\half{\nu}}}{2^{\half{3 \nu}} \Gamma_3(\half{\nu})}  |\cov|^{-\half{\nu + 4}} \exp\left[ -\half{1} \tr{\scl \cov^{-1}} \right].$$ 

The inverse Wishart is the conjugate prior on the covariance matrix of a multivariate Gaussian vector, which means the posterior is also inverse Wishart. Given the data set $\data$ containing $\nobs$ observations and $\scm = \data^\trp \data$ the scatter matrix, the posterior reads: \begin{equation} \label{eqn:posterior}
\cov \given \data \sim \iWish{3}(\scl + \scm, \nu + \nobs). \\
\end{equation}

In order to choose appropriate parameters for the inverse Wishart prior, we analyze the implied distribution in the space of correlation matrices. By transforming the covariance matrix into a correlation matrix, we end up with a so-called \textit{projected inverse Wishart} distribution on the latter, which we denote by $\ipWish$. \cite{barnard_modeling_2000} have shown that if the correlation matrix $\cor$ follows a projected inverse Wishart distribution with scale parameter $\scl$ and $\nu$ degrees of freedom, then the marginal distribution $p(\core_{ij}), i \ne j,$ for off-diagonal elements is uniform if we take $\scl$ to be any diagonal matrix and $\nu = p + 1$, where $p$ is the number of variables. We are working with three variables, so we choose $\nu = 4$.

It is easy to check that for any diagonal matrix $D$, the projected inverse Wishart is scale invariant: 
$$\ipWish(\scl, \nu) \equiv \ipWish(D \scl D, \nu).$$

From~\eqref{eqn:posterior}, it then follows that we can make the posterior distribution on the correlation matrices independent of the scale of the data by choosing the prior scale matrix $\scl = \bfzero_{3, 3}$. Since that would lead to an undefined prior distribution, we can achieve the same goal by setting $\scl = \epsilon \idm_3$ in the limit $\epsilon \downarrow 0$, where $\idm$ is the identity matrix. Summarizing, we will consider the prior distribution: $$\cov \sim \iWish{3}(\epsilon \idm_3, 4), \quad \epsilon \downarrow 0.$$

\subsection{Deriving the Bayes Factors} \label{ssec:derivation}

As reference model ($\model_0$), we choose the most general case in which no independences can be found in the data (`$\var_1 \nindep \var_2 \nindep \var_3$'), which means that the covariance matrix is unconstrained (Figure~\ref{fig:equivalence}, first row). We assume that the covariance matrix follows an inverse Wishart distribution $$p(\cov \given \var_1 \nindep \var_2 \nindep \var_3) = \iWish{3}(\cov; \epsilon \idm_3, \nu),$$ where we consider the limit $\epsilon \downarrow 0$ and set $\nu = 4$ (see Subsection~\ref{ssec:choosing_prior_covm}). Using the conjugacy of the inverse Wishart prior for the covariance matrix, we immediately get the model evidence
\begin{equation} \label{eqn:ref_bf}
	p(\data \given \var_1 \nindep \var_2 \nindep \var_3) = \frac{\epsilon^\frac{3\nu}{2} \Gamma_3(\frac{n + \nu}{2})}{\pi^\frac{3n}{2} \Gamma_3(\frac{\nu}{2})} |\scm + \epsilon \idm_3|^{-\frac{n + \nu}{2}},
\end{equation}
where $\Gamma_p$ is the $p$-variate gamma function and $\scm = \data^\trp \data$ is the scatter matrix. We first compare the evidence for the conditional independence model `$\var_1 \indep \var_2 \indep \var_3$' to the evidence for the reference model `$\var_1 \nindep \var_2 \nindep \var_3$' by computing the Bayes factor:
$$ 	\Bf(\var_1 \indep \var_2 \indep \var_3) = \frac{p(\data \given \var_1 \indep \var_2 \indep \var_3)}{p(\data \given \var_1 \nindep \var_2 \nindep \var_3)}.$$

We can implement the `$\var_1 \indep \var_2 \indep \var_3$' case (Figure~\ref{fig:equivalence}, last row) by constraining $\cov$ to be diagonal, which means we only have to consider the parameters $\cove_{11}, \cove_{22}, \cove_{33}$. We propose to take: 
\begin{equation} \label{eqn:prior_ind}
	p(\cov \given \var_1 \indep \var_2 \indep \var_3) = \prod_{i=1}^3 \iWish{1}(\cove_{ii}; \epsilon, \nu).
\end{equation}

The likelihood also factorizes in this case and becomes
$$ p(\data \given \cov) = \prod_{i=1}^3 \left\{ (2\pi \cove_{ii})^{-\frac{n}{2}} \exp\left[ -\frac{1}{2} \tr{\scme_{ii} \cove^{-1}_{ii}} \right] \right\},$$
yielding the model evidence
$$ p(\data \given \var_1 \indep \var_2 \indep \var_3) =  \prod_{i=1}^3 \left[\frac{\epsilon^\frac{\nu}{2} \Gamma_1(\frac{n + \nu}{2})}{\pi^\frac{\nobs}{2} \Gamma_1(\frac{\nu}{2})} (\scme_{ii} + \epsilon)^{- \frac{\nobs + \nu}{2}}\right].$$

Dividing by the model evidence from~\eqref{eqn:ref_bf} and taking the limit $\epsilon \downarrow 0$, we obtain the Bayes factor%
\begin{equation}
\Bf(\var_1 \indep \var_2 \indep \var_3) = \frac{\Gamma_3(\frac{\nu}{2})}{\Gamma_3(\frac{\nobs + \nu}{2})} \left[\frac{\Gamma_1(\frac{\nobs + \nu}{2})}{\Gamma_1(\frac{\nu}{2})}\right]^3 |\scor|^\frac{\nobs + \nu}{2} = \frac{\nobs + \nu - 2}{\nu - 2} \frac{\Gamma(\frac{\nu - 1}{2})}{\Gamma(\frac{\nobs + \nu}{2})} \frac{\Gamma(\frac{\nu}{2})}{\Gamma(\frac{\nobs + \nu - 1}{2})} |\scor|^\frac{\nobs + \nu}{2}, \label{eqn:bf_indep}
\end{equation}
with $\scor$ the sample correlation matrix and $\Gamma$ the (univariate) gamma function. Due to the choice~\eqref{eqn:prior_ind}, the evidence for `$\var_1 \indep \var_2 \indep \var_3$' also scales with $\epsilon^{\half{\nu}}$, so the dominant terms depending on $\epsilon$ cancel out and the Bayes factor depends only on the correlation matrix in the limit $\epsilon \downarrow 0$. The derivations for the other cases (left out due to space constraints) are similar, leading to the Bayes factors:

\begin{equation}
	\begin{aligned} 
	\Bf(\var_1 \indep \var_2 \indep \var_3) &= f(\nobs, \nu) g(\nobs, \nu) |\scor|^{\frac{\nobs + \nu}{2}} \\
	\Bf(\var_3 \indep (\var_1, \var_2)) &= f(\nobs, \nu) \left(\frac{|\scor|}{1 - \score^2_{12}}\right)^\frac{\nobs + \nu}{2} \\
	\Bf(\var_1 \indep \var_2 \given \var_3) &= g(\nobs, \nu) \left( \frac{|\scor|}{(1 - \score_{13}^2)(1 - \score_{23}^2)} \right)^\frac{\nobs + \nu}{2} \\
	\Bf(\var_1 \indep \var_2) &= \frac{f(\nobs, \nu)}{g(\nobs, \nu)} (1 - \score_{12}^2)^\frac{\nobs + \nu - 1}{2},
	\end{aligned} \label{eqn:bayes_factors}
\end{equation}
where $f(\nobs, \nu) = \dfrac{\nobs + \nu - 2}{\nu - 2}$ and $g(\nobs, \nu) = \dfrac{\Gamma\left(\frac{\nobs + \nu}{2}\right) \Gamma\left(\frac{\nu - 1}{2}\right)}{\Gamma\left(\frac{\nobs + \nu - 1}{2}\right) \Gamma\left(\frac{\nu}{2}\right)} \approx \left(\dfrac{2\nobs + 2\nu - 3}{2 \nu - 3}\right)^\frac{1}{2}$.

In conclusion, for deriving the Bayes factors in~\eqref{eqn:bayes_factors}, we only need to plug in the correlation matrix with the number of samples and compute a limited number of closed-form terms. This is then sufficient to obtain the full posterior distribution over the covariance structures, which makes the BFCS method fast and efficient.

\subsection{Priors on Causal Structures} \label{ssec:prior_cstr}

To do a full Bayesian analysis, we need to specify priors over the different conditional independence models. Assuming faithfulness, there is a one-to-one correspondence between the Markov equivalence classes of the underlying causal graph structure and the conditional independence models. By taking a uniform prior over causal graphs and denoting by $|\model_j|$ the number of causal graphs consistent with the independence model $\model_j \in \mathcal{J}$, we arrive at the prior: $$p(\model_j) = \frac{|\model_j|}{\sum_j |\model_j|}, 
\quad \forall \model_j \in \mathcal{J}.$$

\begin{table}[!htb]
	\centering
	\begin{tabular}{ccc|cccc}
		Pattern & CI Model & Description & DAG & DAG w/ BK & DMAG & DMAG w/ BK \\
		\hline \hline
		Full & $\model_0$ & $\var_1 \nindep \var_2 \nindep \var_3$ & 6 & 2 & 19 & 3 \\ \hline
		& $\model_1$ & $\var_1 \indep \var_2$ & 1 & 1 & 3 & 2 \\
		Acausal & $\model_2$ & $\var_2 \indep \var_3$ & 1 & 0 & 3 & 0 \\
		& $\model_3$ & $\var_3 \indep \var_1$ & 1 & 1 & 3 & 2 \\ \hline
		& $\model_4$ & $\var_1 \indep \var_2 \given \var_3$ & 3 & 1 & 5 & 1 \\
		Causal & $\model_5$ & $\var_2 \indep \var_3 \given \var_1$ & 3 & 1 & 5 & 1 \\
		& $\model_6$ & $\var_3 \indep \var_1 \given \var_2$ & 3 & 1 & 5 & 1 \\ \hline
		& $\model_7$ & $\var_1 \indep (\var_2, \var_3)$ & 2 & 2 & 3 & 3 \\
		Independent& $\model_8$ & $\var_2 \indep (\var_3, \var_1)$ & 2 & 1 & 3 & 1 \\
		& $\model_9$ & $\var_3 \indep (\var_1, \var_2)$ & 2 & 1 & 3 & 1 \\ \hline
		Empty & $\model_{10}$ & $\var_1 \indep \var_2 \indep \var_3$ & 1 & 1 & 1 & 1 \\ \hline \hline
		All & &  & 25 & 12 & 53 & 16 \\
	\end{tabular}
	\caption{Number of causal graph structures over three variables in each Markov equivalence class. In the columns marked `w /BK', the background knowledge that $\var_1$ precedes all other variables, i.e., there can be no arrowhead towards $\var_1$, is added when counting the number of structures.}
	\label{tab:no_str_3var}
\end{table}

In Table~\ref{tab:no_str_3var} we count the number of DAGs and DMAGs (see Section~\ref{sec:bkg}) consistent with each covariance pattern (see Figure~\ref{fig:equivalence}). For example, if we assume an underlying DAG structure, then $p(\var_1 \indep \var_2 \indep \var_3) = \frac{1}{25}$. The addition of background knowledge (BK) reduces the number of causal graph structures corresponding to each conditional independence model. Specifically relevant for discovering causal regulatory relationships is the background knowledge that the genetic marker precedes the expression traits, i.e., that $X_1$ precedes all other variables. This additional constraint leads to the counts in the columns marked `w/ BK' in Table~\ref{tab:no_str_3var}. Some of the covariance patterns imply acausal / causal statements (Figure~\ref{fig:equivalence}), which is what allows us to directly translate the posterior probabilities over covariance patterns into statements over causal relationships.

Now that we have defined priors on the conditional independence models (Markov equivalence classes), we can derive the posterior probabilities from equations~\eqref{eqn:post_ci_model} and~\eqref{eqn:bayes_factors}. The posterior probabilities could also be derived by combining the Bayesian Gaussian equivalent (BGe) score~\citep{geiger_learning_1994} with the priors on causal structures defined in this subsection. Due to our choice of priors on covariance matrices (see Subsection~\ref{ssec:choosing_prior_covm}), however, the Bayes factors are simpler to compute. This makes our approach more efficient when used to infer causal relationships in large regulatory networks. 

To summarize, we have developed a method for computing the posterior probabilities of the covariance structures over three variables (BFCS). We will employ this procedure as part of an algorithm for discovering regulatory relationships. Similarly to LCD and Trigger, the idea is to search over triplets of variables to find potential local causal structures (see Algorithm~\ref{alg:BFCS_yeast}).

\section{Experimental Results} \label{sec:experiments}

\subsection{Consistency of Detecting Local Causal Structures}

\begin{figure}[!htb]
	\begin{subfigure}{.33\linewidth}
		\begin{tikzpicture}[shorten >=1pt, auto, node distance = 2cm, thick]
		\tikzset{vertex/.style = {shape = circle, draw, minimum size=0.5cm}}
		\tikzset{edge/.style = {->,> = latex'}}
		
		\node[vertex] (1) at (0, 0) {$\var_1$};
		\node[vertex] (2) at (2, 0) {$\var_2$};
		\node[vertex] (3) at (4, 0) {$\var_3$};
		
		\draw[edge] (1) to (2);
		\draw[edge] (2) to (3);
		\draw[edge, bend right, draw = none] (1) to (3); 
		\end{tikzpicture}
		\subcaption{Causal model} \label{subfig:generating_causal}
	\end{subfigure}
	\begin{subfigure}{.33\linewidth}
		\begin{tikzpicture}[shorten >=1pt, auto, node distance = 2cm, thick]
		\tikzset{vertex/.style = {shape = circle, draw, minimum size=0.5cm}}
		\tikzset{edge/.style = {->,> = latex'}}
		
		\node[vertex] (1) at (0, 0) {$\var_1$};
		\node[vertex] (2) at (2, 0) {$\var_2$};
		\node[vertex] (3) at (4, 0) {$\var_3$};
		
		\draw[edge] (1) to (2);
		\draw[edge, bend right] (1) to (3);
		\end{tikzpicture}
		\subcaption{Independent model}
	\end{subfigure}
	\begin{subfigure}{.33\linewidth}
		\begin{tikzpicture}[shorten >=1pt, auto, node distance = 2cm, thick]
		\tikzset{vertex/.style = {shape = circle, draw, minimum size=0.5cm}}
		\tikzset{edge/.style = {->,> = latex'}}
		
		\node[vertex] (1) at (0, 0) {$\var_1$};
		\node[vertex] (2) at (2, 0) {$\var_2$};
		\node[vertex] (3) at (4, 0) {$\var_3$};
		
		\draw[edge] (1) to (2);
		\draw[edge] (2) to (3);
		\draw[edge, bend right] (1) to (3);
		\end{tikzpicture}
		\subcaption{Full model}
	\end{subfigure}
	\caption{Generating models} \label{fig:generating_models}
\end{figure}
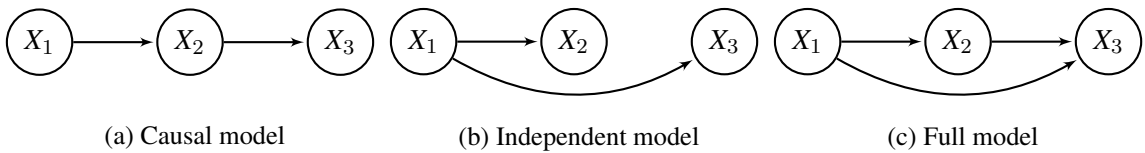

In this simulation, we assessed how well our BFCS approach is able to detect the causal structure $\var_1 \rightarrow \var_2 \rightarrow \var_3$, which is crucial to the application of the LCD and Trigger algorithms. We considered the three generating structures depicted in Figure~\ref{fig:generating_models}. In all three cases, the variables are mutually marginally dependent, but only in the first model $\var_1 \indep \var_3 \given \var_2$ holds.

We sampled random structural parameters (interaction strengths) for each causal relationship from independent standard normal distributions. Then, given each generating model, we generated data sets of different sizes (from $10^2$ up to $10^6$ samples). For each data set, we computed the correlation matrix, which we plugged into~\ref{eqn:bayes_factors} for computing the Bayes factors. We repeated this procedure for 1000 different parameter configurations. We assumed that $\var_1$ precedes all other variables and we allowed for latent variables, so we did not use the knowledge that the data is causally sufficient. We considered a uniform prior over the twelve possible DMAG structures (see Table~\ref{tab:no_str_3var}). 

\begin{figure}[!htb]
	\begin{subfigure}{.33\linewidth}
		\includegraphics[width = \textwidth]{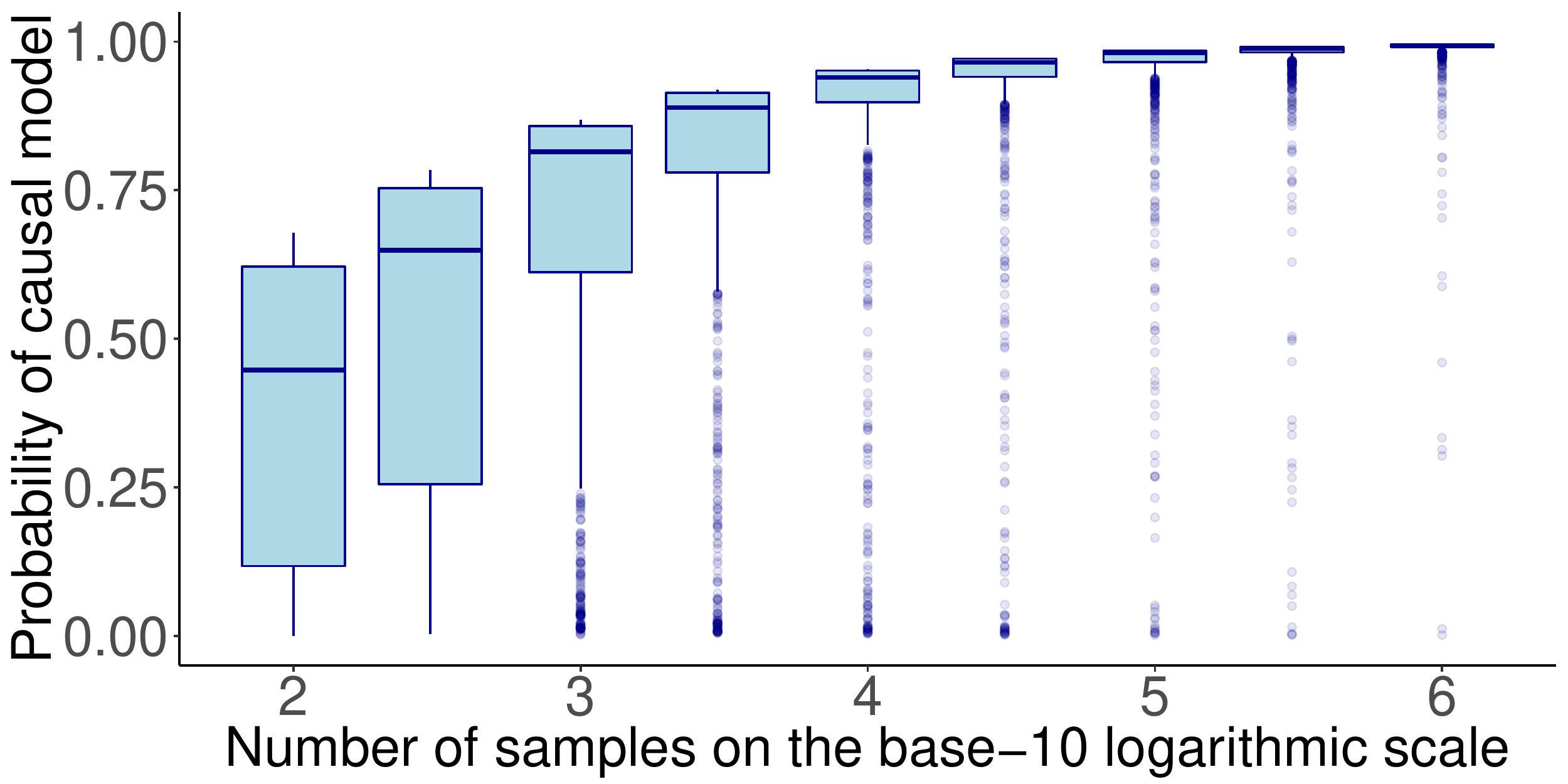}
		\subcaption{Causal model}
		\label{subfig:consistency_T_norm}
	\end{subfigure}
	\begin{subfigure}{.33\linewidth}
		\includegraphics[width = \textwidth]{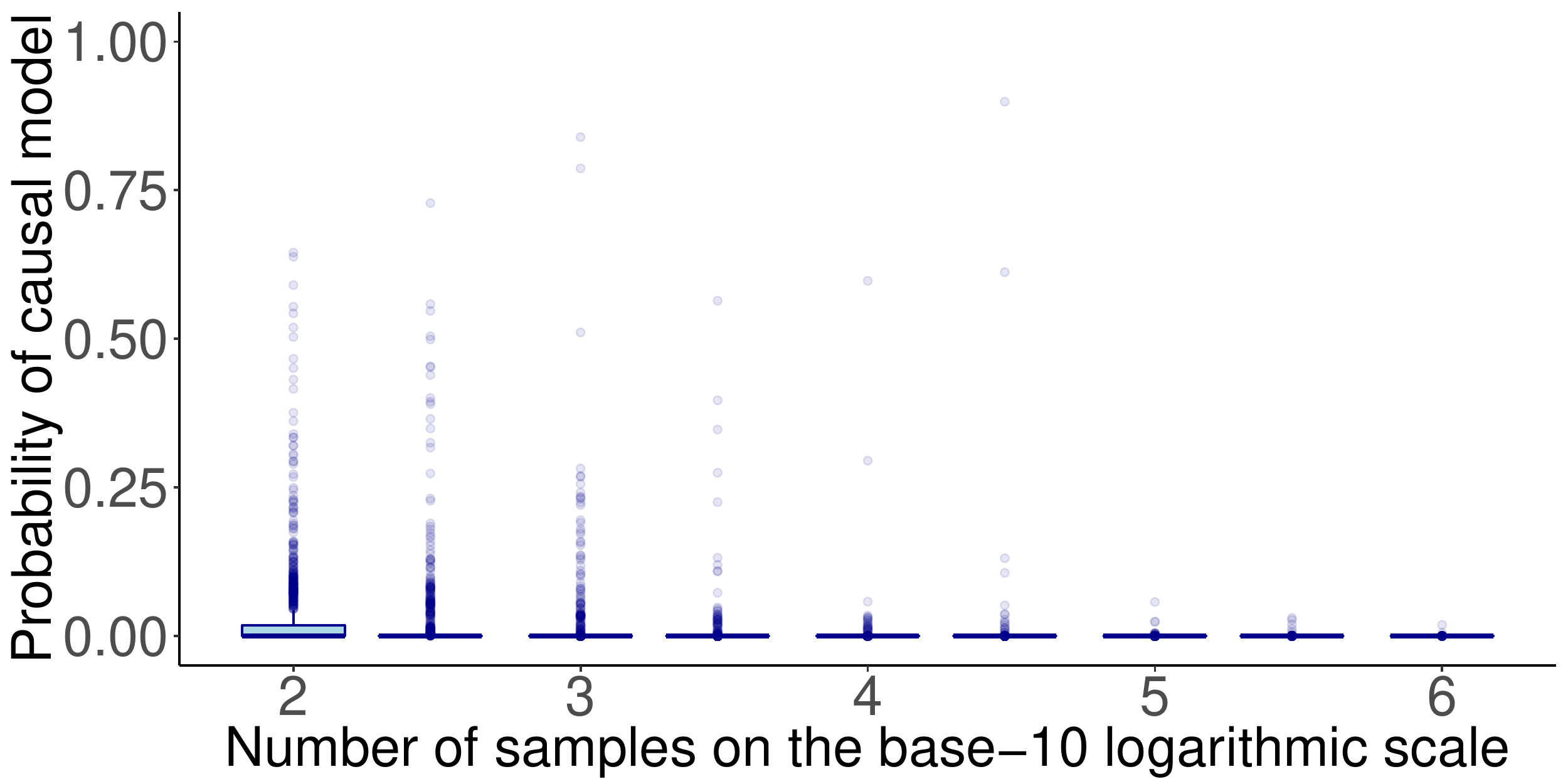}
		\subcaption{Independent model}
	\end{subfigure}
	\begin{subfigure}{.33\linewidth}
		\includegraphics[width = \textwidth]{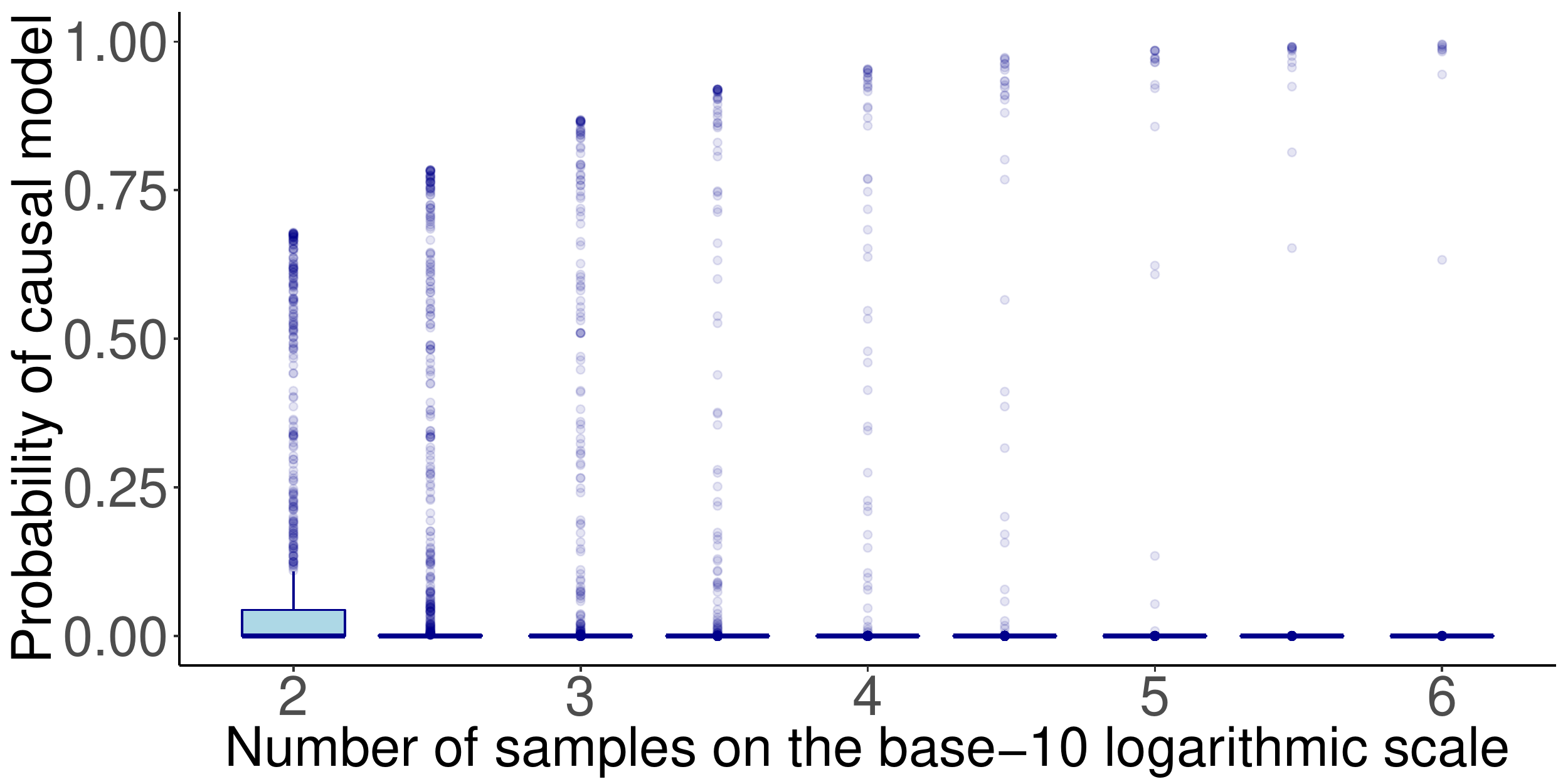}
		\subcaption{Full model}
	\end{subfigure}
	\begin{subfigure}{.33\linewidth}
		\includegraphics[width = \textwidth]{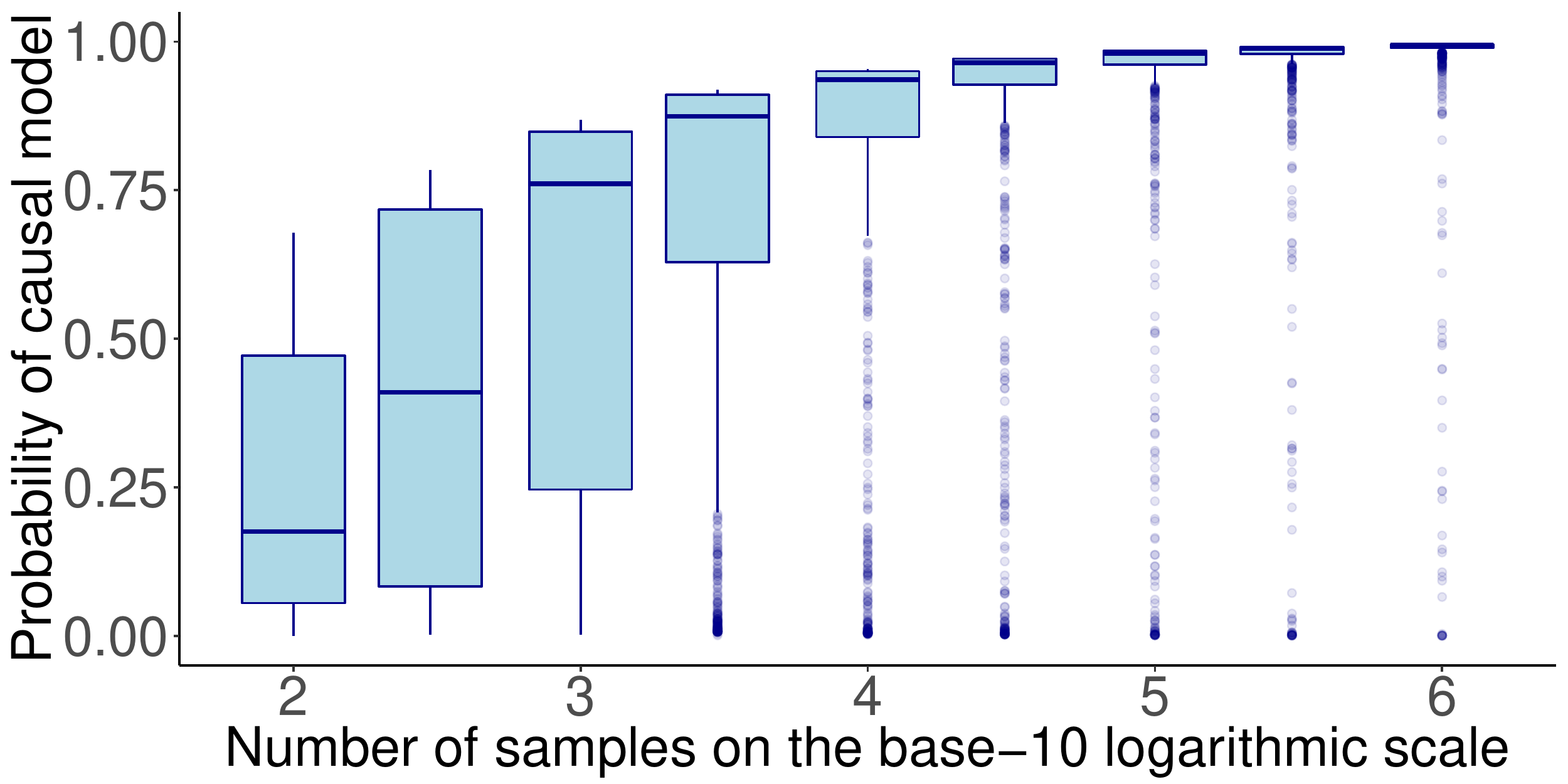}
		\subcaption{Causal model} \label{subfig:consistency_T_binom}
	\end{subfigure}
	\begin{subfigure}{.33\linewidth}
		\includegraphics[width = \textwidth]{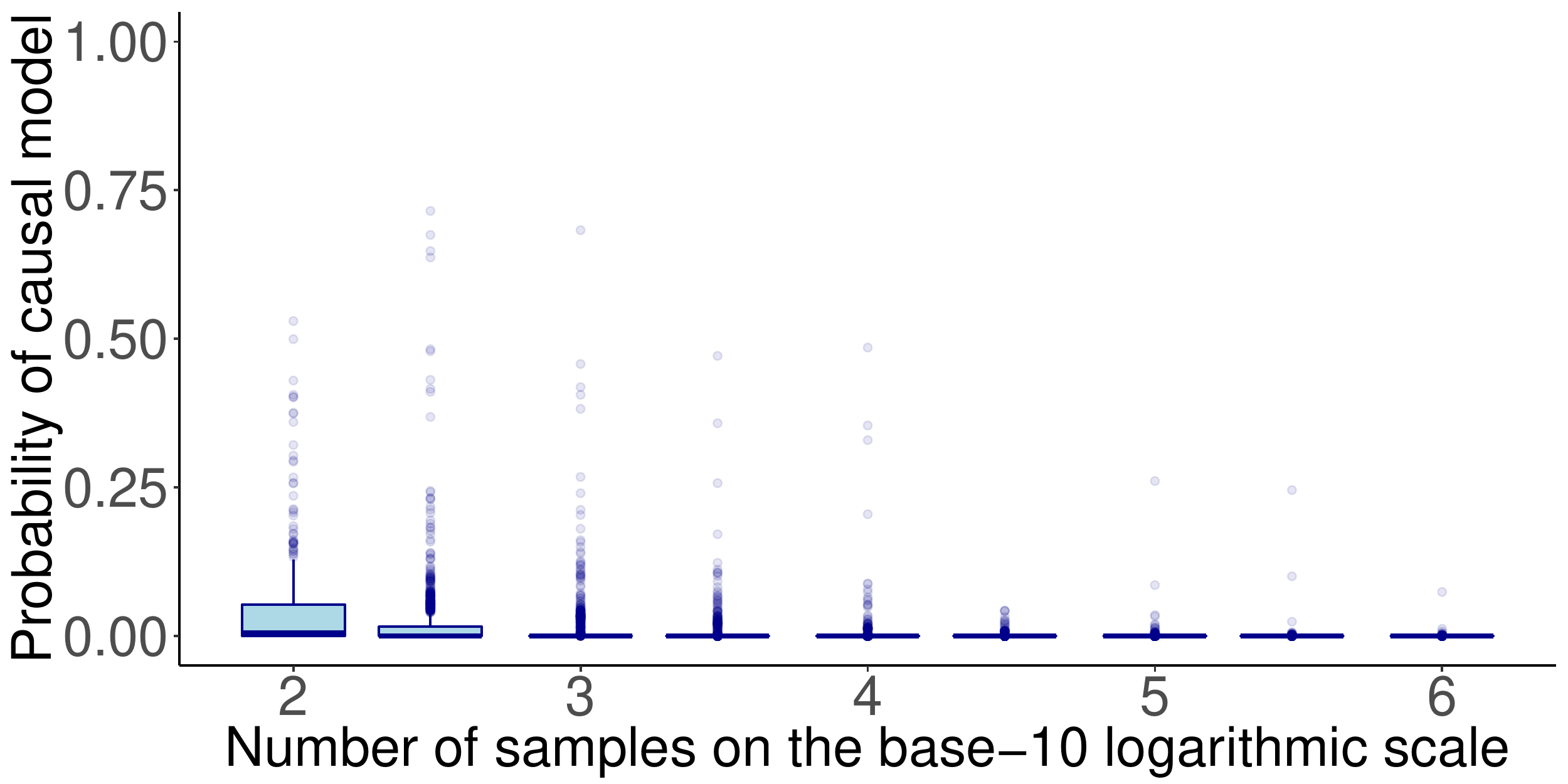}
		\subcaption{Independent model}
	\end{subfigure}
	\begin{subfigure}{.33\linewidth}
		\includegraphics[width = \textwidth]{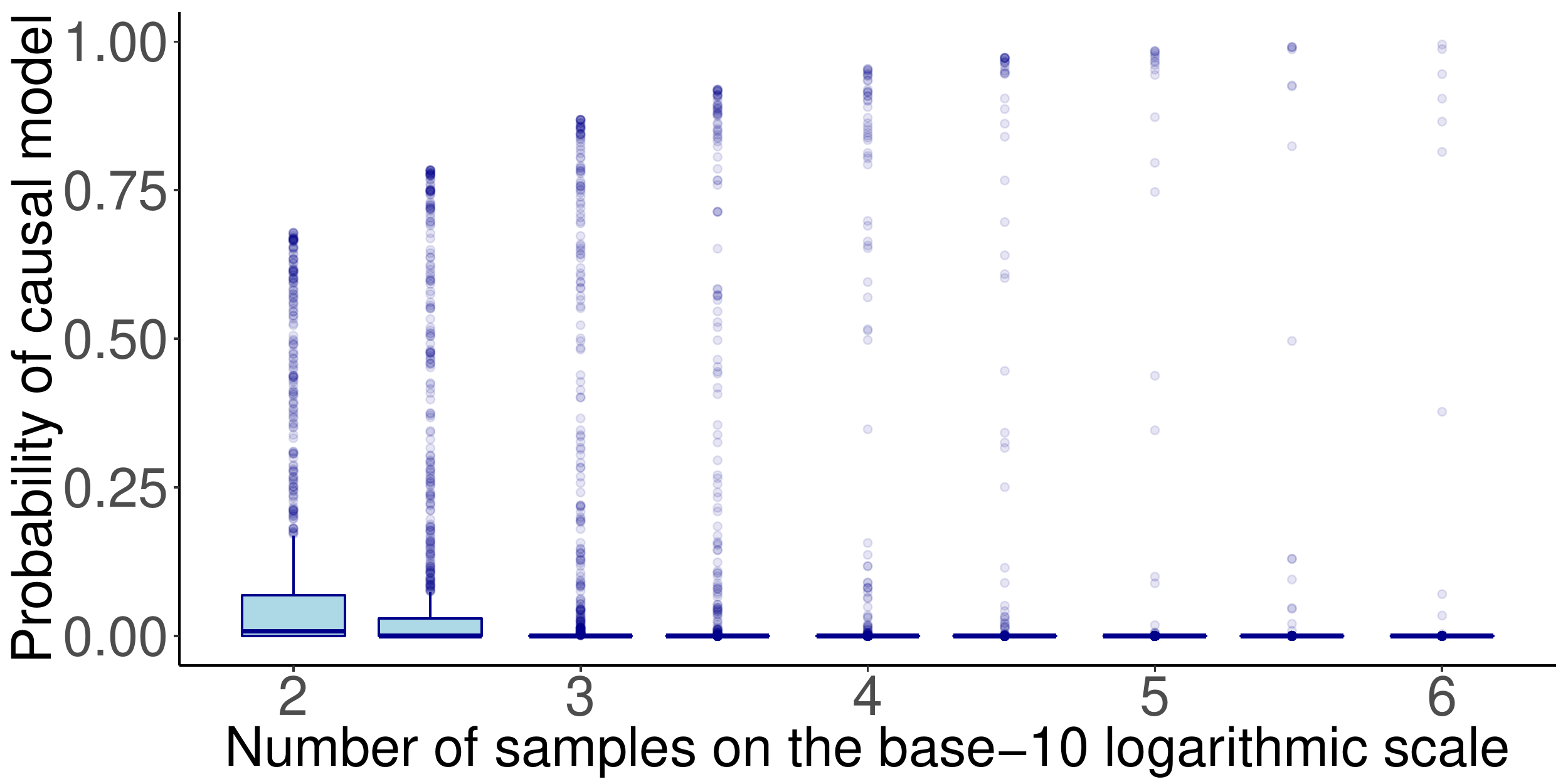}
		\subcaption{Full model}
	\end{subfigure}
	\caption{Box plots of the posterior probabilities of $p(\var_1 \rightarrow \var_2 \rightarrow \var_3 \given \data)$ output by BFCS across the 1000 different parameter configurations for each of the generating models in Figure~\ref{fig:generating_models}. As we increased the number of samples, the probability $p(\var_1 \rightarrow \var_2 \rightarrow \var_3 \given \data)$ converged to one when the data was generated from the causal model (a) and converged to zero when it was not (b and c). \textbf{Top}: $\varv = (\var_1, \var_2, \var_3)$ is multivariate Gaussian; \textbf{Bottom}: $\var_1$ is Bernoulli and $(\var_2, \var_3) \given \var_1$ is Gaussian.} \label{fig:consistency}
\end{figure}

In the first experiment we generated random multivariate data from the models in Figure~\ref{fig:generating_models}. As expected, the posterior probability $p(\var_1 \rightarrow \var_2 \rightarrow \var_3 \given \data)$ converged to one (Figure~\ref{fig:consistency}, top row) when the true generating model was the one in Figure~\ref{subfig:generating_causal}. At the same time, $p(\var_1 \rightarrow \var_2 \rightarrow \var_3 \given \data)$ converged to zero when the true generating model was the independent or full model. Note that it is easier to distinguish the causal model from the independent model than from the full model. When generating data from the full model, it is possible to generate structural parameters that are close to zero. If the direct interaction between $\var_1$ and $\var_3$ is close to zero, it looks as if $\var_1 \indep \var_3 \given \var_2$.

In the second experiment, we considered the same generating models, but we sampled $\var_1$ from a Bernoulli distribution to mimic a genetic variable, e.g., an allele or the parental strain in the yeast experiment~\citep{chen_harnessing_2007}. We sampled a different success probability for the Bernoulli variable in each repetition from a uniform distribution between 0.1 and 0.9. The random vector $(\var_2, \var_3)$ was sampled from a bivariate Gaussian conditional on the value of $\var_1$. In the bottom row of Figure~\ref{fig:consistency} we see that when the Gaussian assumption did not hold for $\var_1$, we lost some power in recovering the correct model. When we increased the number of samples, this loss in power due to the violation of the Gaussian assumption became less severe and BFCS remained consistent.

\subsection{Causal Discovery in Gene Regulatory Networks}

\begin{figure}[!htb]
	\centering
	\includegraphics[width=\linewidth]{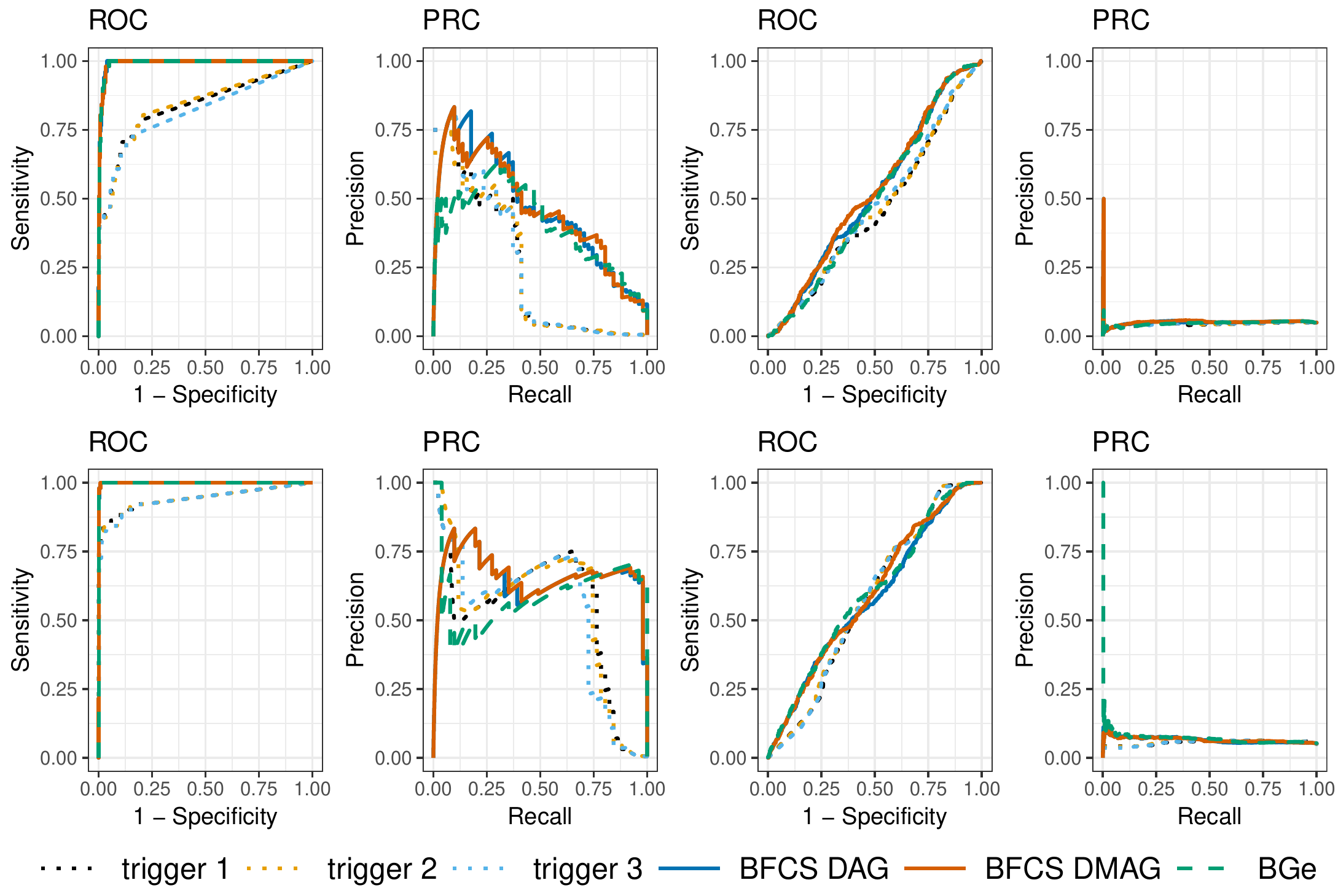}
	\caption{Comparing the performance of Trigger and BFCS in terms of the ROC and Precision-Recall (PRC) curves. We generated both a sparse GRN (\textbf{left column}) and a dense GRN (\textbf{right column}) consisting of 51 and 491 regulatory relationships, respectively. We generated 100 samples (\textbf{top row}) and then 1000 samples (\textbf{bottom row}) from both networks. We ran Trigger three times (labeled `trigger 1', `trigger 2', and `trigger 3') on the simulated data to account for differences when sampling the null statistics. We compared Trigger against two versions of BFCS, in which we took a uniform prior over DAGs (`BFCS DAG') and DMAGs (`BFCS DMAG'), respectively. For reference, we also show the performance of an equivalent method that uses the BGe score (`BGe').}
	\label{fig:prroc}
\end{figure}

In this experiment, we simulated a transcriptional regulatory network meant to emulate the yeast data set analyzed in ~\cite{chen_harnessing_2007}. We randomly generated data for 100 genetic markers, where each marker is an independent Bernoulli variable with `success' probability uniformly sampled between 0.1 and 0.5. We then generated transcript level expression data from the structural equation model:
$$ \textbf{t} := \textbf{B} \textbf{t} + \textbf{l} + \bm{\varepsilon},$$
where $\textbf{B}$ is a lower triangular matrix, $\textbf{t} = \left(T_1, T_2, ..., T_{100}\right)^\trp$ is the random vector of the expression trait data, $\textbf{l} = \left(L_1, L_2, ..., L_{100}\right)^\trp$ is the random vector of the genetic markers, and $\bm{\varepsilon} \sim \mathcal{N}(\bfzero_{100}, \idm_{100})$ is added noise. The true causal relationships are in the directed graph structure defined by $\textbf{B}$. Each expression trait $T_i$ was causally linked to the genetic marker $L_i$, $\forall i \in \{1, 2, ..., 100\}$.

For each pair of expression traits $(T_i, T_j)$ and for every genetic marker $L_k$, we derived the posterior probability of $L_k \rightarrow T_i \rightarrow T_j$ using BFCS. We then took the maximum of these probabilities over $k$, which is a lower bound of the probability that there exists $L_k$ such that the triple $(L_k, T_i, T_j)$ has the causal structure $L_k \rightarrow T_i \rightarrow T_j$. Similarly to Trigger, we reported this value as a conservative estimate for the probability of $T_i \rightarrow T_j$. For reference, we also evaluated the performance of an equivalent method that uses the BGe score instead of the Bayes factors computed in BFCS.  

In Figure~\ref{fig:prroc}, we compare the performance of Trigger and BFCS by showing the ROC and Precision-Recall (PRC) curves. Both Trigger and BFCS performed much better when the underlying network structure was sparse. For sparse networks, BFCS shows a significant improvement in the AUC measure for both curves. For dense networks, Trigger and BFCS have fairly similar ROC curves, but BFCS shows a noticeable improvement in the precision-recall curve. More specifically, the higher precision for the first causal relationships that are recalled shows that BFCS is better at ranking the top regulatory relationships. Furthermore, the probabilities output by BFCS are better calibrated than those by Trigger (see Table~\ref{tab:calibration}). This is because Trigger is overconfident in its predictions, which explains the large spread in probabilities compared to BFCS.

\begin{table}[!htb]
	\centering
	\footnotesize
	\begin{tabular}{c||cccc|cccc}
		& \multicolumn{4}{c|}{Sparse GRN} & \multicolumn{4}{c}{Dense GRN} \\
		Samples & trigger & BFCS DAG & BFCS DMAG & BGe & trigger & BFCS DAG & BFCS DMAG & BGe \\ \hline
		100 & 0.037 & 0.022 & 0.015 & 0.009 & 0.481 & 0.201 & 0.148 & 0.231 \\
		1000 & 0.028 & 0.011 & 0.008 & 0.006 & 0.343 & 0.376 & 0.310 & 0.445 \\
	\end{tabular}
	\caption{Comparing the calibration of Trigger and BFCS using the Brier score (lower is better).} 
	\label{tab:calibration}
\end{table}

\subsection{Comparing Results from an Experiment on Yeast}

\begin{algorithm}[!htb]
	\caption{Running BFCS on the yeast data set} 
	\label{alg:BFCS_yeast}
	\begin{algorithmic}[1] 
		\State \textit{Input:} Yeast data set consisting of 3244 markers and 6216 gene expression measurements
		\ForAll{expression traits $T_i$}
			\ForAll{expression traits $T_j, j \neq i$}
				\ForAll{genetic markers $L_k$}
					\State Compute the Bayes factors for the triplet $(L_k, T_i, T_j)$
					\State Derive the posterior probability of the structure $L_k \rightarrow T_i \rightarrow T_j$ given the data
				\EndFor
				\State Save $\max_k p(L_k \rightarrow T_i \rightarrow T_j)$ as the probability of gene $i$ regulating gene $j$
			\EndFor
		\EndFor
		\State \textit{Output:} Matrix of regulation probabilities 
	\end{algorithmic}
\end{algorithm}

\cite{chen_harnessing_2007} showcased the Trigger algorithm by applying it to an experiment on yeast, in which two distinct strains were crossed to produce 112 independent recombinant segregant lines. Genome-wide genotyping and expression profiling were performed on each segregant line. We computed the probabilities over the triples in the yeast data set using \texttt{trigger}~\citep{chen_trigger_2017} and BFCS (Algorithm~\ref{alg:BFCS_yeast}). For BFCS, we used DMAGs to allow for the possibility of latent variables. 

\begin{table}[!htb]
	\footnotesize \centering
	\begin{subtable}{.49\linewidth}
		\begin{tabular}{c|cccc}
			Rank & Gene & Chen et al. & trigger & BFCS \\ \hline
			1 & MDM35 & 0.973 & 0.999 & 0.678 \\ 
			2 & CBP6 & 0.968 & 0.997 & 0.683 \\ 
			3 & QRI5 & 0.960 & 0.985 & 0.678 \\ 
			4 & RSM18 & 0.959 & 0.984 & 0.672 \\ 
			5 & RSM7 & 0.953 & 0.977 &  0.684 \\ 
			6 & MRPL11 & 0.924 & 0.999 & 0.670 \\ 
			7 & MRPL25 & 0.887 & 0.908 & 0.675 \\ 
			8 & DLD2 & 0.871 & 0.896 &  0.660 \\ 
			9 & YPR126C & 0.860 & 0.904 &  0.634 \\ 
			10 & MSS116 & 0.849 & 0.997 & 0.659 \\ 
		\end{tabular}%
		\subcaption{Genes regulated by NAM9,  sorted by `Chen et al.'.}
		\label{tab:NAM9_reg_chen} 
	\end{subtable}
	\begin{subtable}{.49\linewidth}
		\begin{tabular}{c|cccc}
			Rank & Gene & Chen et al. & trigger & BFCS \\ \hline
			1 & FMP39 & 0.176 & 0.401 & 0.691 \\ 
			2 & DIA4 & 0.493 & 0.987 & 0.691 \\ 
			3 & MRP4 & 0.099 & 0.260 & 0.691 \\ 
			4 & MNP1 & 0.473 & 0.999 & 0.691 \\ 
			5 & MRPS18 & 0.527 & 0.974 & 0.690 \\ 
			6 & MTG2 & 0.000 & 0.000 & 0.690 \\ 
			7 & YNL184C & 0.299 & 0.768 & 0.690 \\ 
			8 & YPL073C & 0.535 & 0.993 & 0.690 \\ 
			9 & MBA1 & 0.290 & 0.591 & 0.690 \\ 
			10 & ACN9 & 0.578 & 0.927 & 0.690 \\ 
		\end{tabular}
		\subcaption{Genes regulated by NAM9, sorted by `BFCS'. }
		\label{tab:NAM9_reg_bfcs} 
	\end{subtable}

	\caption{ The column `Chen et al.' shows the original results of the Trigger algorithm as reported in~\cite{chen_harnessing_2007}. The `trigger' column contains the probabilities we obtained when running the algorithm from the Bioconductor \texttt{trigger} package ~\citep{chen_trigger_2017} on the entire yeast data set with default parameters. The column `BFCS' contains the output of running Algorithm~\ref{alg:BFCS_yeast} on the yeast data set, for which we took a uniform prior over DMAGs.}
	\label{tab:NAM9_reg} 
\end{table}

In Table~\ref{tab:NAM9_reg_chen}, we report the top ten genes purported to be regulated by the putative regulator NAM9, sorted according to the probability estimates reported in~\cite{chen_harnessing_2007}. We see that BFCS also assigns relatively high, albeit better calibrated (Table~\ref{tab:calibration}) and much more conservative, probabilities to the most significant regulatory relationships found by Trigger. In Table~\ref{tab:NAM9_reg_bfcs}, we see that some relationships ranked significant by BFCS are assigned a very small probability by Trigger. The regulatory relationship NAM9 $\rightarrow$ MTG2, of which both genes are part of the mitochondrial ribosome assembly, is ranked sixth by BFCS, but is assigned zero probability by Trigger. This is because, in the Trigger algorithm, the genetic marker $L_i$ exhibiting the strongest linkage with $T_i$ (NAM9 in this case) is preselected and then only the probability of $L_i \rightarrow T_i \rightarrow T_j$ is estimated. With BFCS, on the other hand, we estimated the probability of this structure for all genetic markers.

\section{Discussion} \label{sec:discussion}

We have introduced a novel Bayesian approach for inferring gene regulatory networks that uses the information in the local covariance structure over triplets of variables to make statements about the presence of causal relationships. One key advantage of our method is that we consider all possible causal structures at once, whereas other methods only look at and test for a subset of structures. Because we focus on discovering local causal structures, our method is simple, fast, and inherently parallel, which makes it applicable to very large data sets. Furthermore, the probability estimates produced by BFCS constitute a measure of reliability in the inferred causal relations. We have demonstrated the effectiveness of our approach by comparing it against the Trigger algorithm, a state-of-the-art procedure for inferring causal regulatory relationships. Other methods for inferring gene regulatory networks such as `CIT'~\citep{millstein_disentangling_2009} or `CMST'~\citep{neto_modeling_2013} output $p$-values instead of probability estimates, which is why they are not directly comparable to BFCS.

In this paper, we have proposed a simple uniform prior on two types of causal graph structures, namely DAGs and DMAGs. However, our method allows for more informative causal priors to be incorporated, taking into consideration properties such as the sparsity of the networks. Moreover, our approach is structure-agnostic, by which we mean we can consider different causal graph structures incorporating various data-generating assumptions. The tricky part is then to come up with an appropriate prior on the set of causal graph structures from which we assume the data is generated. 
 

\acks{This research has been partially financed by the Netherlands Organisation for Scientific Research (NWO), under project 617.001.451. We would like to thank the anonymous reviewers for their thoughtful comments and efforts towards improving our manuscript.}

\vskip 0.2in
\bibliography{bucur18a-arxiv}

\begin{thebibliography}{13}
\providecommand{\natexlab}[1]{#1}
\providecommand{\url}[1]{\texttt{#1}}
\expandafter\ifx\csname urlstyle\endcsname\relax
  \providecommand{\doi}[1]{doi: #1}\else
  \providecommand{\doi}{doi: \begingroup \urlstyle{rm}\Url}\fi

\bibitem[Barnard et~al.(2000)Barnard, McCulloch, and
  Meng]{barnard_modeling_2000}
J.~Barnard, R.~McCulloch, and X.-L. Meng.
\newblock Modeling {Covariance} {Matrices} in {Terms} of {Standard}
  {Deviations} and {Correlations}, with {Application} to {Shrinkage}.
\newblock \emph{Stat. Sin.}, 10\penalty0 (4):\penalty0 1281--1311, 2000.

\bibitem[Chen et~al.(2007)Chen, Emmert-Streib, and
  Storey]{chen_harnessing_2007}
L.~S. Chen, F.~Emmert-Streib, and J.~D. Storey.
\newblock Harnessing naturally randomized transcription to infer regulatory
  relationships among genes.
\newblock \emph{Genome Biology}, 8:\penalty0 R219, Oct. 2007.

\bibitem[Chen et~al.(2017)Chen, Sangurdekar, and Storey]{chen_trigger_2017}
L.~S. Chen, D.~P. Sangurdekar, and J.~D. Storey.
\newblock trigger, 2017.
\newblock URL \url{https://doi.org/doi:10.18129/B9.bioc.trigger}.

\bibitem[Claassen and Heskes(2011)]{claassen_logical_2011}
T.~Claassen and T.~Heskes.
\newblock A {Logical} {Characterization} of {Constraint}-based {Causal}
  {Discovery}.
\newblock {UAI}'11, pages 135--144, Arlington, Virginia, United States, 2011.
  AUAI Press.

\bibitem[Cooper(1997)]{cooper_simple_1997}
G.~F. Cooper.
\newblock A {Simple} {Constraint}-{Based} {Algorithm} for {Efficiently}
  {Mining} {Observational} {Databases} for {Causal} {Relationships}.
\newblock \emph{Data Mining and Knowledge Discovery}, 1\penalty0 (2):\penalty0
  203--224, June 1997.

\bibitem[Geiger and Heckerman(1994)]{geiger_learning_1994}
D.~Geiger and D.~Heckerman.
\newblock Learning {{Gaussian Networks}}.
\newblock UAI'94, pages 235--243, San Francisco, CA, USA, 1994. {Morgan
  Kaufmann Publishers Inc.}
\newblock ISBN 978-1-55860-332-5.

\bibitem[Mani et~al.(2006)Mani, Spirtes, and Cooper]{mani_theoretical_2006}
S.~Mani, P.~Spirtes, and G.~F. Cooper.
\newblock A {Theoretical} {Study} of {Y} {Structures} for {Causal} {Discovery}.
\newblock {UAI}'06, pages 314--323, Arlington, Virginia, United States, 2006.
  AUAI Press.

\bibitem[Meinshausen and
  Bühlmann(2006)]{nicolai_meinshausen_high-dimensional_2006}
N.~Meinshausen and P.~Bühlmann.
\newblock High-{Dimensional} {Graphs} and {Variable} {Selection} with the
  {Lasso}.
\newblock \emph{The Annals of Statistics}, 34\penalty0 (3):\penalty0
  1436--1462, 2006.

\bibitem[Millstein et~al.(2009)Millstein, Zhang, Zhu, and
  Schadt]{millstein_disentangling_2009}
J.~Millstein, B.~Zhang, J.~Zhu, and E.~E. Schadt.
\newblock Disentangling molecular relationships with a causal inference test.
\newblock \emph{BMC Genetics}, 10:\penalty0 23, May 2009.

\bibitem[Neto et~al.(2013)Neto, Broman, Keller, Attie, Zhang, Zhu, and
  Yandell]{neto_modeling_2013}
E.~C. Neto, A.~T. Broman, M.~P. Keller, A.~D. Attie, B.~Zhang, J.~Zhu, and
  B.~S. Yandell.
\newblock Modeling {Causality} for {Pairs} of {Phenotypes} in {System}
  {Genetics}.
\newblock \emph{Genetics}, 193\penalty0 (3):\penalty0 1003--1013, Mar. 2013.

\bibitem[Richardson and Spirtes(2002)]{richardson_ancestral_2002}
T.~Richardson and P.~Spirtes.
\newblock Ancestral {Graph} {Markov} {Models}.
\newblock \emph{The Annals of Statistics}, 30\penalty0 (4):\penalty0 962--1030,
  2002.
\newblock ISSN 0090-5364.

\bibitem[Sadeghi and Lauritzen(2014)]{sadeghi_markov_2014}
K.~Sadeghi and S.~Lauritzen.
\newblock Markov properties for mixed graphs.
\newblock \emph{Bernoulli}, 20\penalty0 (2):\penalty0 676--696, May 2014.
\newblock ISSN 1350-7265.
\newblock \doi{10.3150/12-BEJ502}.

\bibitem[Studeny(2006)]{studeny_probabilistic_2006}
M.~Studeny.
\newblock \emph{Probabilistic {Conditional} {Independence} {Structures}}.
\newblock Springer Science \& Business Media, June 2006.
\newblock ISBN 978-1-84628-083-2.
\newblock Google-Books-ID: NJ4iwCMoznIC.

\end{thebibliography}

\end{document}